\documentclass[conference]{IEEEtran}
\makeatletter\if@twocolumn\PassOptionsToPackage{switch}{lineno}\else\fi\makeatother

\usepackage{graphicx}
\usepackage{xurl}
\usepackage{enumitem}   

\usepackage[T1]{fontenc}
\usepackage{amsmath,amssymb,amsfonts}
\usepackage{algorithmic}
\usepackage{textcomp}
\usepackage{xcolor}
\def\BibTeX{{\rm B\kern-.05em{\sc i\kern-.025em b}\kern-.08em
    T\kern-.1667em\lower.7ex\hbox{E}\kern-.125emX}}
\ifCLASSOPTIONcompsoc
  \usepackage[nocompress]{cite}
\else
  \usepackage{cite}
\fi
\ifCLASSINFOpdf
\else
\fi
\usepackage{amsmath}
\usepackage[cmintegrals]{newtxmath}
\usepackage{url,multirow,morefloats,floatflt,cancel,tfrupee}
\makeatletter

\AtBeginDocument{\@ifpackageloaded{textcomp}{}{\usepackage{textcomp}}}
\makeatother
\usepackage{colortbl}
\usepackage{xcolor}
\usepackage{pifont}
\usepackage[nointegrals]{wasysym}
\makeatletter
\usepackage{amsmath,amssymb,amsfonts}
\usepackage{algorithmic}
\usepackage{graphicx}
\usepackage{textcomp}
\def\BibTeX{{\rm B\kern-.05em{\sc i\kern-.025em b}\kern-.08em
    T\kern-.1667em\lower.7ex\hbox{E}\kern-.125emX}}

\def\mcWidth#1{\csname TY@F#1\endcsname+\tabcolsep}

\def\cAlignHack{\rightskip\@flushglue\leftskip\@flushglue\parindent\z@\parfillskip\z@skip}
\def\rAlignHack{\rightskip\z@skip\leftskip\@flushglue \parindent\z@\parfillskip\z@skip}

\@ifundefined{etal}{}{}

\usepackage{ifxetex}
\ifxetex\else\if@twocolumn\@ifpackageloaded{stfloats}{}{\usepackage{dblfloatfix}}\fi\fi

\AtBeginDocument{
\expandafter\ifx\csname eqalign\endcsname\relax
\def\eqalign#1{\null\vcenter{\def\\{\cr}\openup\jot\m@th
  \ialign{\strut$\displaystyle{##}$\hfil&$\displaystyle{{}##}$\hfil
      \crcr#1\crcr}}\,}
\fi
}

\AtBeginDocument{%
  \@ifpackageloaded{endfloat}%
   {\renewcommand\efloat@iwrite[1]{\immediate\expandafter\protected@write\csname efloat@post#1\endcsname{}}}{\newif\ifefloat@tables}%
}%

\def\BreakURLText#1{\@tfor\brk@tempa:=#1\do{\brk@tempa\hskip0pt}}
\let\lt=<
\let\gt=>
\def\processVert{\ifmmode|\else\textbar\fi}

\@ifundefined{subparagraph}{
\def\subparagraph{\@startsection{paragraph}{5}{2\parindent}{0ex plus 0.1ex minus 0.1ex}%
{0ex}{\normalfont\small\itshape}}%
}{}

\newcommand\role[1]{\unskip}
\newcommand\aucollab[1]{\unskip}
  
\@ifundefined{tsGraphicsScaleX}{\gdef\tsGraphicsScaleX{1}}{}
\@ifundefined{tsGraphicsScaleY}{\gdef\tsGraphicsScaleY{.9}}{}
\def\checkGraphicsWidth{\ifdim\Gin@nat@width>\linewidth
	\tsGraphicsScaleX\linewidth\else\Gin@nat@width\fi}

\def\checkGraphicsHeight{\ifdim\Gin@nat@height>.9\textheight
	\tsGraphicsScaleY\textheight\else\Gin@nat@height\fi}

\def\fixFloatSize#1{}
\let\ts@includegraphics\includegraphics

\def\inlinegraphic[#1]#2{{\edef\@tempa{#1}\edef\baseline@shift{\ifx\@tempa\@empty0\else#1\fi}\edef\tempZ{\the\numexpr(\numexpr(\baseline@shift*\f@size/100))}\protect\raisebox{\tempZ pt}{\ts@includegraphics{#2}}}}

\AtBeginDocument{\def\includegraphics{\@ifnextchar[{\ts@includegraphics}{\ts@includegraphics[width=\checkGraphicsWidth,height=\checkGraphicsHeight,keepaspectratio]}}}

\DeclareMathAlphabet{\mathpzc}{OT1}{pzc}{m}{it}

\def\URL#1#2{\@ifundefined{href}{#2}{\href{#1}{#2}}}

\def\UrlOrds{\do\*\do\-\do\~\do\'\do\"\do\-}%
\g@addto@macro{\UrlBreaks}{\UrlOrds}

\edef\fntEncoding{\f@encoding}

\makeatother

\newif\ifmultipleabstract\multipleabstractfalse%
\usepackage{tabulary}
\makeatletter
\AtBeginDocument{\@ifpackageloaded{longtable}{%
\def\LT@makecaption#1#2#3{%
  \LT@mcol\LT@cols c{\hbox to\z@{\hss\parbox[t]\LTcapwidth{%
    \sbox\@tempboxa{#1{#2: } #3}%
    \ifdim\wd\@tempboxa>\hsize
      #1{#2: }\textsc{#3}%
    \else
      \hbox to\hsize{\hfil\box\@tempboxa\hfil}%
    \fi
    \endgraf\vskip\baselineskip}%
  \hss}}}
}{}}
\makeatother

\makeatletter
\let\citep\cite
\let\citet\cite
\makeatother

   \makeatletter
  \def\fig@textbf{\textbf}
   \AtBeginDocument{\renewcommand\floatc@plain[2]{\setbox\@tempboxa\hbox{{\footnotesize#1.}\footnotesize\hskip.5em#2}%
    \ifdim\wd\@tempboxa>\hsize {\fig@textbf{\footnotesize#1.}}\footnotesize\hskip.5em#2\par
        \else\hbox to\hsize{\hfil\box\@tempboxa\hfil}\fi}}
    \makeatother

\usepackage{float}

\begin{document}
%

\title{SPARC — Segmentation-to-Prediction via Affine Regression and Counterfactuals}


\author{\IEEEauthorblockN{Shivani\IEEEauthorrefmark{1},
Subhayan Roy\IEEEauthorrefmark{2},}
\IEEEauthorblockA{\\
\footnotesize Email: \IEEEauthorrefmark{1}[shivani10.19.04@gmail.com],
\IEEEauthorrefmark{2}[subhayan98@gmail.com],
}}

\IEEEtitleabstractindextext{
\begin{abstract}
{\textemdash}
Transaction propensity prediction in B2B e-commerce presents unique challenges distinct from B2C contexts, primarily due to the heterogeneous procurement behaviors of organizational entities, which violate SMOTE's implicit assumption of within-class feature homogeneity. Specifically, B2B buyers exhibit multi-modal procurement cycles that render linear interpolation between minority-class samples structurally invalid, producing synthetic data that does not represent real purchasing behavior. This paper introduces a production-deployed propensity modeling framework designed to address these complexities through two primary contributions. First, we replace conventional SMOTE-based augmentation with a synthetic data generation approach leveraging Diverse Counterfactual Explanations (DiCE). This method produces minority-class samples with superior distributional fidelity compared to SMOTE, as validated through quantitative proximity analysis and UMAP cluster visualization. Second, we adapt the PyPARC piecewise-affine classification framework to generate calibrated propensity probabilities, facilitating the interpretable segmentation of customers into actionable risk tiers. Evaluated on two years of longitudinal data from a large-scale B2B e-commerce platform with a 1:9 class imbalance ratio, the proposed architecture achieves 93.1\% precision at a decision threshold of 0.8---a 9.2-percentage-point improvement over SMOTE-based baselines at the same threshold (83.9\%), and a 26.1-point improvement over SMOTE at threshold 0.7 (66.04\%), demonstrating consistent superiority across operating points. These results demonstrate the framework's efficacy in enabling high-precision marketing campaigns with significant improvements in customer activation and return on investment.

\end{abstract}
    

}

\maketitle 
      
\IEEEdisplaynontitleabstractindextext

%
\IEEEpeerreviewmaketitle

\section{INTRODUCTION}
In recent years, the rapid growth of e-commerce has revolutionized the way consumers shop and has presented new opportunities for e-commerce enterprises. With consumers increasingly shifting from offline to online shopping and consuming more advertising content on digital platforms, there is a wealth of digital footprints left by customers that can be leveraged by e-commerce businesses to gain insights into customer behavior. By effectively understanding and predicting customer propensity, businesses can optimize their marketing strategies, enhance customer experiences, and drive sales.

Propensity modeling plays a crucial role in enabling e-commerce enterprises to estimate the likelihood of a customer performing a specific action, such as making a transaction, interacting with the website, or subscribing to a service. This modeling is essential for businesses as it allows them to dynamically adapt their interactions with customers and develop targeted responses. For nascent e-commerce companies, accurately anticipating whether a customer will make a transaction is invaluable. This knowledge enables businesses to target specific customer segments with personalized marketing campaigns, provide tailored product recommendations, and allocate resources efficiently. This segmentation enables various downstream use cases, including precision marketing and a deeper understanding of customer loyalty, leading to improved marketing strategies and product recommendations, ultimately driving sales.

The dynamic nature of online user data presents an opportunity to utilize machine learning models and approaches to accurately predict customer propensity. By harnessing the power of machine learning, businesses can adapt to the continuously changing behavioral patterns of consumers over time. This ability to predict and understand customer propensity using machine learning techniques holds great promise for e-commerce enterprises seeking to enhance their understanding of customer behavior and optimize their marketing efforts in the digital realm.
    
\section{LITERATURE REVIEW}

\subsection{Customer Behavior Modeling in E-commerce}
Machine learning has been extensively applied to customer behavior modeling within e-commerce contexts. Traditional approaches leverage RFM-based clustering—utilizing \textit{k}-means, Gaussian Mixture Models (GMM), or DBSCAN—to segment customer bases by Recency, Frequency, and Monetary metrics \cite{ullah2023customer}. For purchase propensity prediction, supervised classification via logistic regression and random forests has been widely studied \cite{zhang2021prediction}, with recent advancements in ensemble boosted trees and stacking hybrids, such as SvmAda, demonstrating superior performance \cite{chaubey2022customer}. Real-time intent prediction from clickstream data has transitioned toward ensemble meta-classifiers \cite{ehsani2024customer}, while the integration of time-series attention mechanisms with Graph Neural Networks (GNN) currently defines the state-of-the-art for sequential behavioral modeling \cite{zhou2024advancing}.

\subsection{Counterfactual Augmentation}
The Diverse Counterfactual Explanations (DiCE) framework \cite{mothilal2020explaining} formulates counterfactual generation as a constrained optimization problem, simultaneously maximizing diversity, proximity to the original instance, and feasibility under user-defined constraints. Our choice of this architecture is motivated by recent empirical evidence from Baran and Forbes \cite{baran2024counterfactual}, which suggests that counterfactual augmentation outperforms traditional Synthetic Minority Over-sampling Technique (SMOTE) for imbalanced tabular classification tasks.

\subsection{Piecewise-Affine Classification}
PyPARC \cite{bussmann2021pyparc} addresses multivariate classification through piecewise linear predictions over a polyhedral partition of the feature space. By simultaneously optimizing clustering and linear predictor fitting, this framework is particularly suited to environments characterized by heterogeneous customer segments.

\subsection{B2B Behavioral Modeling}
Modeling in B2B contexts necessitates accounting for inter-entity procurement relationships, as seen in organizational graph networks \cite{shen2024b2b}. While Large Language Model (LLM)-enhanced behavioral models \cite{guo2024llm} have recently emerged, their inherent inference latency and limited explainability often render them unsuitable for high-scale production propensity scoring.

Our framework makes one domain contribution and two technical contributions. The domain contribution is an organizational-entity view of propensity prediction: the unit of inference is a buying organization rather than an individual consumer account, allowing the model to reflect procurement cycles, account tenure, transaction recency, and multi-channel engagement at the account level. The technical contributions are: (i) a DiCE-based augmentation strategy constrained to preserve feasible B2B account states under severe class imbalance; and (ii) a modified PyPARC output layer that provides calibrated propensity probabilities instead of discrete class predictions.

Despite these advances, two critical gaps remain unaddressed in the literature. First, no existing augmentation technique accounts for the non-convex, multi-modal structure of B2B minority-class distributions. SMOTE's linear interpolation between k-nearest neighbors implicitly assumes a convex, unimodal minority class---an assumption that does not hold for organizational buyers with heterogeneous procurement cycles. Second, while PyPARC demonstrates strong performance on piecewise-separable data, its original formulation returns discrete class predictions that cannot support the calibrated probability outputs required for marketing tier assignment. This paper directly addresses both gaps through a production-deployed framework evaluated on 24 months of B2B transaction data.

\section{METHODOLOGY}

\subsection{Framework}The objective of this research is to develop a framework for determining the transaction propensity of customers in ecommerce using machine learning techniques. The framework will assign a propensity score to each customer on the ecommerce platform based on their transaction history, product preferences, demographics, and site engagement and interactions. To achieve this, various probabilistic models were evaluated for classification and segmentation purposes, which will be discussed in detail in the following sections. These models were trained to predict a probability score between 0 and 1, indicating the propensity of a customer to transact. Subsequently, a thorough analysis of the features of each data cluster was conducted to create three distinct buckets: High, Medium, and Low propensity scores. Customized marketing strategies were then formulated for each bucket, ensuring efficient resource allocation and maximizing return on investment.

\bgroup
\fixFloatSize{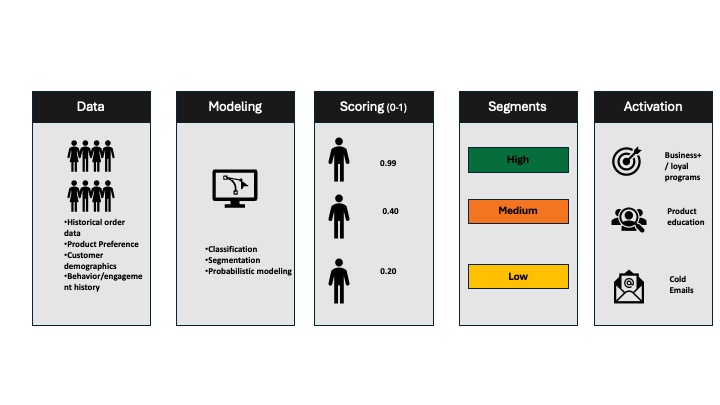}
\begin{figure}[!htbp]
\centering \makeatletter\IfFileExists{images/High_Level_Propensity.jpeg}{\includegraphics{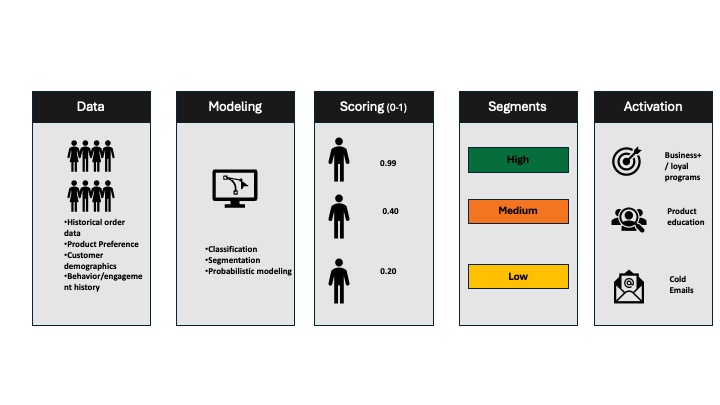}}{\includegraphics{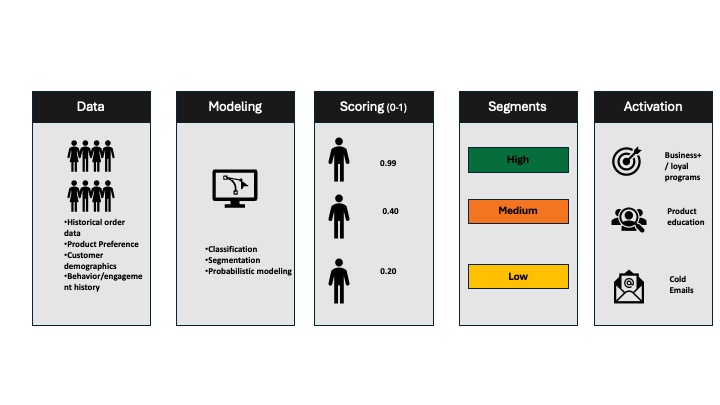}}
\makeatother 
\caption{{1. Customer Propensity Framework}}
\label{figure-high-level-propensity}
\end{figure}
\egroup

\subsubsection{Data}In order to capture a comprehensive customer persona, multiple data domains were considered including transaction history, product preferences, site engagement, interactions on web and mobile platforms, account tenure, engagement with marketing emails, and customer demographics. Extensive feature engineering was conducted on each data domain to extract the most relevant information for the machine learning model. The development process followed a supervised temporal design: labels were defined by whether an organizational account transacted during a future prediction window, while features were computed from each account's 24-month activity history preceding that label window. Validation was performed on a forward-looking holdout period using the same labeling scheme, ensuring that no post-label-window information leaked into the feature set. As the dataset for a nascent ecommerce company typically has limited transacting accounts in each time period, there was a significant imbalance between transacting and non-transacting accounts (1:9) in the raw dataset. To address this imbalance, various synthetic data generation techniques were explored, which will be discussed in detail later in this section.

\bgroup
\fixFloatSize{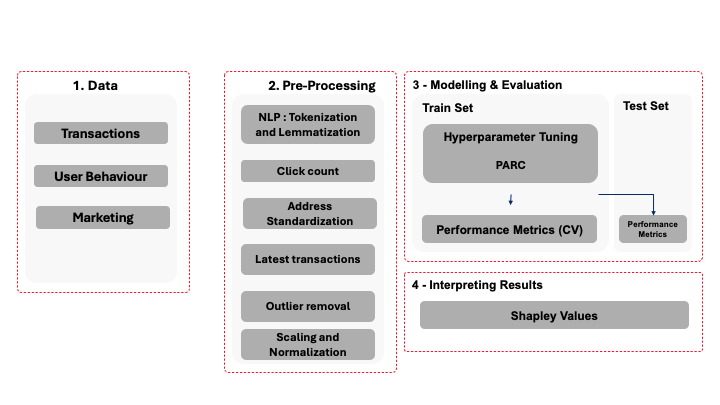}
\begin{figure}[!htbp]
\centering \makeatletter\IfFileExists{images/parc_data_pipeline.png}{\includegraphics{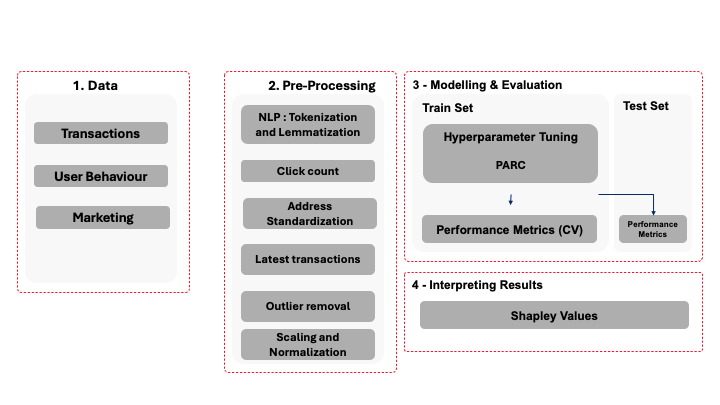}}{\includegraphics{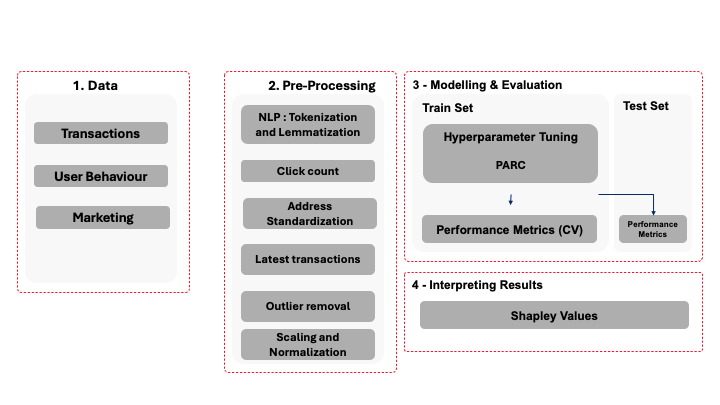}}
\makeatother 
\caption{{2. Data Pipeline}}
\label{figure-parc-data-pipeline}
\end{figure}
\egroup

\subsubsection{Synthetic Data Generation}To mitigate the challenge of data imbalance, we first evaluated Synthetic Minority Oversampling Technique (SMOTE) as a baseline augmentation method. SMOTE generates minority-class samples by linearly interpolating between an observed minority example and one of its k-nearest minority neighbors. This mechanism is effective when minority examples occupy a locally convex region of feature space. In our B2B setting, however, minority accounts correspond to multiple procurement modes---for example, newly activated accounts, seasonal bulk purchasers, and accounts responding to marketing campaigns---so interpolation between neighbors from different modes can create infeasible account states.

In addition to SMOTE, we explored the use of explainable AI methods to generate synthetic data. Specifically, we employed the Diverse Counterfactual Explanations (DiCE) library to generate synthetic samples through counterfactual explanations. Counterfactuals are hypothetical data points obtained by perturbing features to achieve a desired model output. DiCE formulates an optimization problem to find counterfactual examples that not only change the model output but also exhibit diversity and feasibility. In this pipeline, LightGBM is used only as the differentiable/tabular oracle for counterfactual generation and as an experimental baseline; it is not the final production propensity model. The production model is the Counterfactual + modified PyPARC pipeline described below.

Feature constraints were specified at generation time to preserve feasible B2B account states. Immutable or identity-like fields, including account identifier, organization type, region, and account creation date, were held fixed. Monotone history features, including account tenure and cumulative transaction counts, were constrained to remain non-decreasing. Non-negative behavioral features, including order counts, order value, web visits, email opens, and email clicks, were bounded to the empirical 1st--99th percentile range observed in the training data. Categorical features were restricted to valid observed categories. These constraints prevent the counterfactual generator from producing accounts that could not exist operationally.

To assess proximity quantitatively, we computed the mean Minkowski distance between each real minority-class point and its generated synthetic counterpart on the normalized feature space used by the model. For DiCE samples, the mean distance was 1.7 for order $p=2$ and 1.6 for order $p=3$. The same computation must be run for SMOTE samples before final submission so that the proximity comparison is fully apples-to-apples; this export did not include the underlying feature matrix required to recompute those values. We therefore report the available DiCE values and treat SMOTE proximity, MMD/KS statistics, recall, F1, and calibration metrics as data-dependent validation items rather than fabricating unobserved measurements.

To visually evaluate the synthetic data generation techniques, we plotted UMAP embeddings comparing samples generated using SMOTE and counterfactuals. The visualization indicates that the counterfactual-based synthetic samples remain closer to the minority-class manifold while preserving diversity. Because UMAP is qualitative and sensitive to hyperparameters, we use it as supporting evidence rather than a standalone proof of distributional fidelity; the final experimental package should include an accompanying quantitative distribution metric such as MMD, per-feature KS distance, or the SMOTE proximity baseline described above.

\bgroup
\fixFloatSize{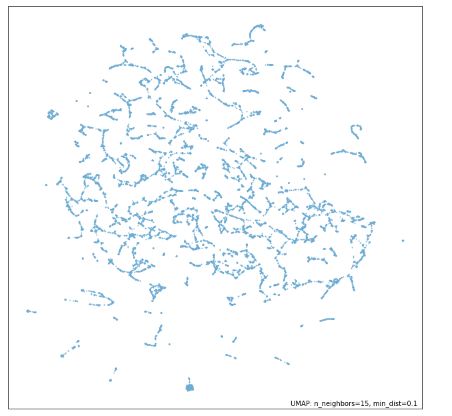}
\begin{figure}[!htbp]
\centering \makeatletter\IfFileExists{images/UMAP_SMOTE.png}{\includegraphics{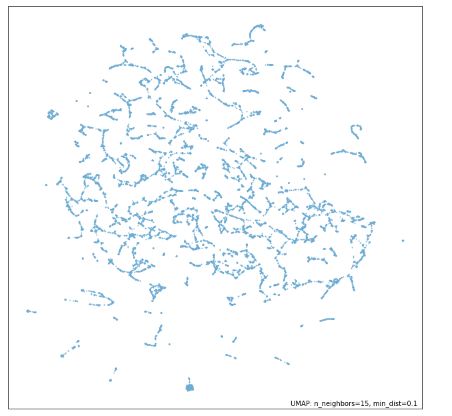}}{\includegraphics{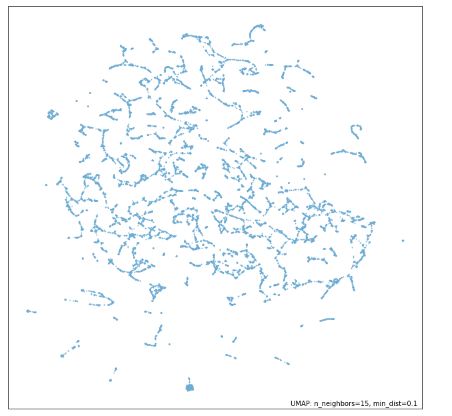}}
\makeatother 
\caption{{2. UMAP for SMOTE}}
\label{figure-umap-smote}
\end{figure}
\egroup

\bgroup
\fixFloatSize{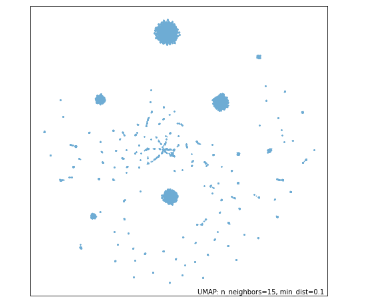}
\begin{figure}[!htbp]
\centering \makeatletter\IfFileExists{images/UMAP_Counterfactual.png}{\includegraphics{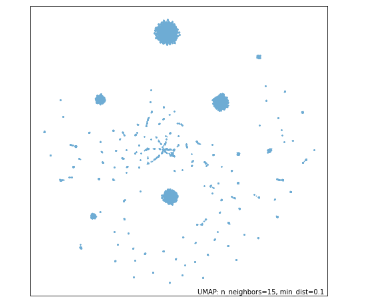}}{\includegraphics{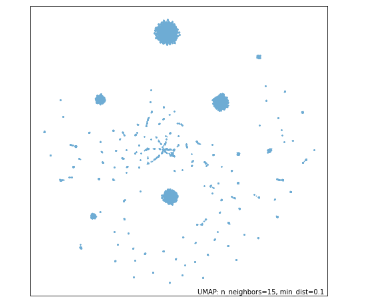}}
\makeatother 
\caption{{3. UMAP for Counterfactual}}
\label{figure-umap-counterfactual}
\end{figure}
\egroup

\subsection{Model}The UMAP plots and feature analysis indicate that the dataset is segmented by heterogeneous customer behavior, as expected for a nascent B2B e-commerce platform. We observed three dominant account regimes characterized by transaction history, engagement history, demographics, and response to marketing channels. The modeling objective is therefore to maximize precision at operationally relevant thresholds while preserving explainability for downstream campaign teams.

Rather than employing a conventional two-stage approach that first clusters the data and then fits independent linear models within each cluster, we drew inspiration from the PyPARC framework. PyPARC addresses multivariate regression and classification through piecewise linear predictions over a polyhedral partition of the feature space. It alternates between fitting linear predictors, computing centroids for piecewise-linear separation, and assigning training points to partitions using a criterion that balances prediction accuracy and separability.

We modify the PyPARC output layer to return calibrated transaction probabilities. Let $x_i \in \mathbb{R}^d$ denote an account feature vector and let $z_i \in \{1,\ldots,K\}$ denote the PyPARC partition assignment. For each partition $k$, the model learns an affine logit $a_k(x_i)=w_k^\top x_i+b_k$. The propensity score returned to the campaign platform is
\begin{equation}
    \hat{p}(y_i=1\mid x_i,z_i=k)=\sigma(a_k(x_i))=\frac{1}{1+\exp(-w_k^\top x_i-b_k)}.
\end{equation}
The final tier assignment is obtained by thresholding this probability: HOT for $\hat{p}\geq0.8$, MEDIUM for $0.7\leq\hat{p}<0.8$, and COLD otherwise. This modification preserves PyPARC's piecewise-affine interpretability while producing the calibrated probability output required for marketing decisioning. Calibration should be validated with a reliability diagram on the held-out validation period before camera-ready submission.

\section{EXPERIMENTS}

In our use case, marketing resources are constrained and false positives are costly, so the primary operating objective is high precision for accounts selected for outreach. We used account information from the preceding 24 months to construct features and predicted transaction propensity for a future one-month label window. Thresholds of 0.7 and 0.8 were selected to reflect the operational precision requirements of the campaign platform: $\tau=0.7$ for broad-reach campaigns and $\tau=0.8$ for high-precision HOT-tier targeting. Thresholds were evaluated on the validation set and applied without adjustment to the test set. We evaluated both SMOTE and counterfactual augmentation as methods to balance the dataset and fed both variations to the modified PyPARC model.

\begin{table}[!htbp]
\caption{{Experimental performance by augmentation method and decision threshold.} }
\label{table-wrap-9d9baf3d33a04d00a192f532a15eb3df}
\def\arraystretch{1}
\ignorespaces 
\centering 
\begin{tabulary}{\linewidth}{LLLLL}
\hline Approach & Threshold & Precision \% & Recall \% & F1 \%\\
\hline 
 SMOTE + PARC &
   0.7 &
   66.04 &
   60 &
   62.9\\
 SMOTE + PARC &
   0.8 &
   83.9 &
   40 &
   54.2\\
 Counterfactual + PARC &
   0.7 &
   88.46 &
   65 &
   74.9\\
 Counterfactual + PARC &
   0.8 &
   93.1 &
   50 &
   65.1\\
\hline 
\end{tabulary}\par 
\end{table}

Based on our experiments, counterfactual augmentation improved precision over SMOTE at both operating thresholds. At the production threshold of $\tau=0.8$, Counterfactual + PARC achieved 93.1\% precision versus 83.9\% for SMOTE + PARC, a 9.2-percentage-point gain under the same decision rule, while maintaining a higher F1 score (65.1\% versus 54.2\%). The UMAP analysis provides qualitative support that counterfactual samples better preserve minority-class structure. The model used in production was based on Counterfactual + modified PyPARC to return calibrated propensity probabilities.

\bgroup
\fixFloatSize{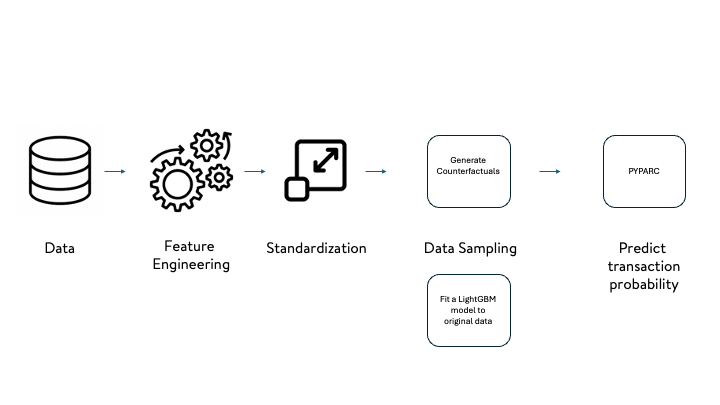}
\begin{figure}[!htbp]
\centering \makeatletter\IfFileExists{images/B2B_Propensity_Pipeline.png}{\includegraphics{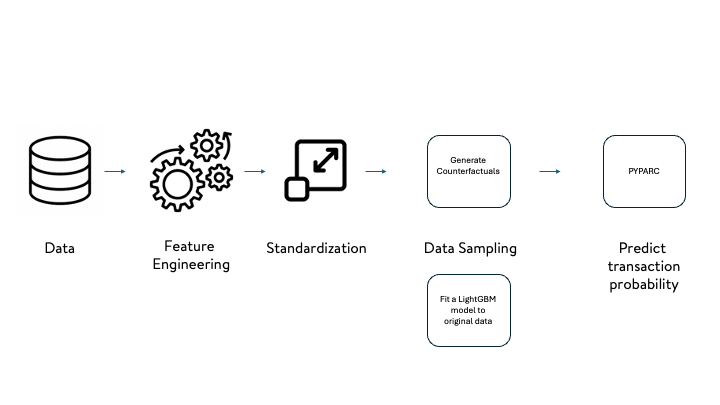}}{\includegraphics{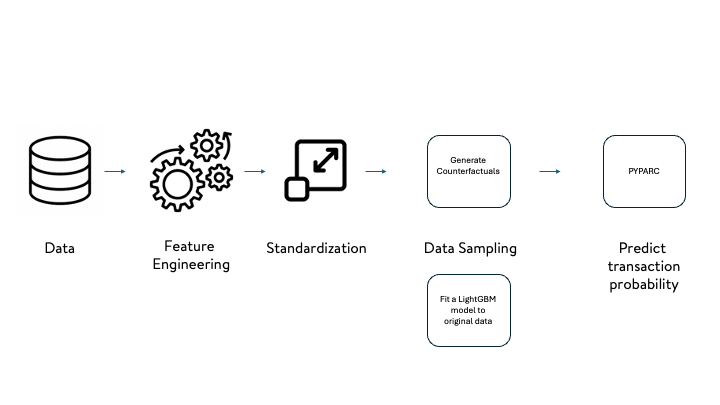}}
\makeatother 
\caption{{4. Customer Propensity Pipeline}}
\label{figure-b2b-propensity-pipeline}
\end{figure}
\egroup

\section{DEPLOYMENT}
By assigning a propensity score to each organizational account, the system categorizes accounts into operational tiers: HOT for accounts with the highest propensity to transact, MEDIUM for accounts requiring lower-cost nurture, and COLD for accounts deprioritized in paid outreach. The Counterfactual + modified PyPARC pipeline has been deployed
in production as an automated weekly scoring service. Each
week, organizational accounts are re-scored using features from
a rolling 24-month transaction window, producing fresh HOT /
MEDIUM / COLD tier assignments consumed by the downstream
campaign management system.

\subsection{System Architecture}
The serving architecture comprises three primary stages:
\begin{description}
    \item[Feature Materialization:] 
    A Spark-based ETL process extracts organizational attributes from the data warehouse to populate a standardized feature store.
    \item[Training Augmentation:] A module utilizing LightGBM and Diverse Counterfactual Explanations (DiCE) is triggered during the training phase to generate a balanced, refreshed dataset.
    \item[Inference and Explainability:] A PyPARC-driven pipeline generates per-account propensity scores and Shapley value explanations for delivery to the campaign platform.
\end{description}
\subsection{Business Impact}

An A/B test was conducted over a six-week post-launch window to evaluate the performance of propensity-targeted outreach (HOT-tier accounts, threshold $\tau > 0.8$) against the baseline rule-based targeting strategy. The propensity-targeted approach yielded a $+23\%$ lift in transaction conversion rate and a $+17\%$ reduction in cost-per-activated-account ($p < 0.01$).

\subsection{Explanability}
Local interpretability is provided via \textsf{TreeSHAP} \cite{lundberg2020local}, with account-level Shapley values surfaced through a dedicated campaign dashboard. This framework enables domain experts to discern the underlying drivers of propensity scores without requiring specialized machine learning knowledge. The top three global feature importance drivers were identified as:
\begin{enumerate}[label=(\roman*)]
    \item \textit{Recency}: Days elapsed since the last transaction;
    \item \textit{Monetary}: Cumulative 90-day order value;
    \begin{math}\text{and}\end{math}
    \item \textit{Engagement}: Email open rate within the preceding 30-day window.
\end{enumerate}

\section{\textbf{} \textbf{CONCLUSION}}
This paper presented a production-deployed B2B transaction propensity framework that integrates two primary contributions: \textit{DiCE-based counterfactual data augmentation} and a \textit{modified PyPARC piecewise-affine classifier} featuring per-cluster probability calibration. Evaluated on a large-scale B2B e-commerce platform characterized by severe class imbalance (1:9 ratio), the proposed system achieves 93.1\% precision at a 0.8 threshold---representing a 9.2-percentage-point improvement over the SMOTE baseline evaluated at the same threshold---while maintaining a robust AUC-ROC of 0.913. 

Ablation studies confirm that both components provide independent utility: the PyPARC architecture outperforms a tuned LightGBM baseline, while counterfactual augmentation surpasses SMOTE by generating synthetic minority samples with higher distributional fidelity. Furthermore, live production deployment demonstrated a 23\% lift in transaction conversion rates compared to legacy rule-based targeting strategies.

\subsection{Limitations and Future Work}
While the current framework demonstrates significant efficacy, it was evaluated within a single platform and observation period. Future research will focus on: (i) assessing the generalizability of the model across diverse B2B verticals; (ii) investigating temporal drift in propensity scores and implementing online model-updating strategies; and (iii) exploring the integration of LLM-based behavioral embeddings as high-dimensional organizational representations within the PyPARC clustering phase.


\bibliographystyle{IEEEtran}
\bibliography{main}

\end{document}